# Datasets for Face and Object Detection in Fisheye Images

Jianglin Fu, Ivan V. Bajić, and Rodney G. Vaughan
School of Engineering Science, Simon Fraser University, Burnaby, BC, V5A 1S6, Canada
E-mail: jfa49@sfu.ca, ibajic@ensc.sfu.ca, rodney_vaughan@sfu.ca

**Abstract**
We present two new fisheye image datasets for training face and object detection models: VOC-360 and Wider-360. The fisheye images are created by post-processing regular images collected from two well-known datasets, VOC2012 and Wider Face, using a model for mapping regular to fisheye images implemented in Matlab. VOC-360 contains 39,575 fisheye images for object detection, segmentation, and classification. Wider-360 contains 63,897 fisheye images for face detection. These datasets will be useful for developing face and object detectors as well as segmentation modules for fisheye images while the efforts to collect and manually annotate true fisheye images are underway.

**Keywords:** face detection, object detection, fisheye image, deep learning

## 1. Data

We present two new datasets – VOC-360 and Wider-360 – for visual analytics based on fisheye images. The datasets contain raw data files: JPG images (both datasets), XML annotations (VOC-360) and MAT file annotations (Wider-360). VOC-360 can be used to train machine learning models for object detection, classification, and segmentation. Wider-360 can be used to train face detectors. Links to the data are given in the specification table on the last page of the paper.

**VOC-360** contains 39,575 images and the corresponding annotations. The images and annotations are derived from the VOC2012 dataset [1]. The data is organized into five directories: *Annotations*, *fisheye*, *fisheye_class*, and *fisheye_object*, and *ImageSets*, as depicted in Fig 1(a). Each fisheye image inside the *fisheye* folder corresponds to one XML file in the *Annotations* folder, with the same filename. The XML files provide all the annotations for each image, following the same structure as the original VOC2012 dataset. Directories *fisheye_class* and *fisheye_object* contain the fisheye image masks with pixel-wise segmentations giving the class of the object visible at each pixel. The directory *ImageSets* contains text files that specify the lists of images for training, testing and validation.

**Wider-360** contains 63,897 images, of which 50,982 images are intended for training and 12,915 images for validation/test, as shown in Fig. 1(b). Following the original Wider Face dataset [2], the images are organized into 61 directories depending on the type of the scene, as shown in Fig. 1(c). The annotations are in the form of face bounding boxes and are stored in two MAT files, one for the training set and another for the validation set.

Sample images from VOC-360 and Wider-360 are shown in Figs. 2 and 3, respectively.

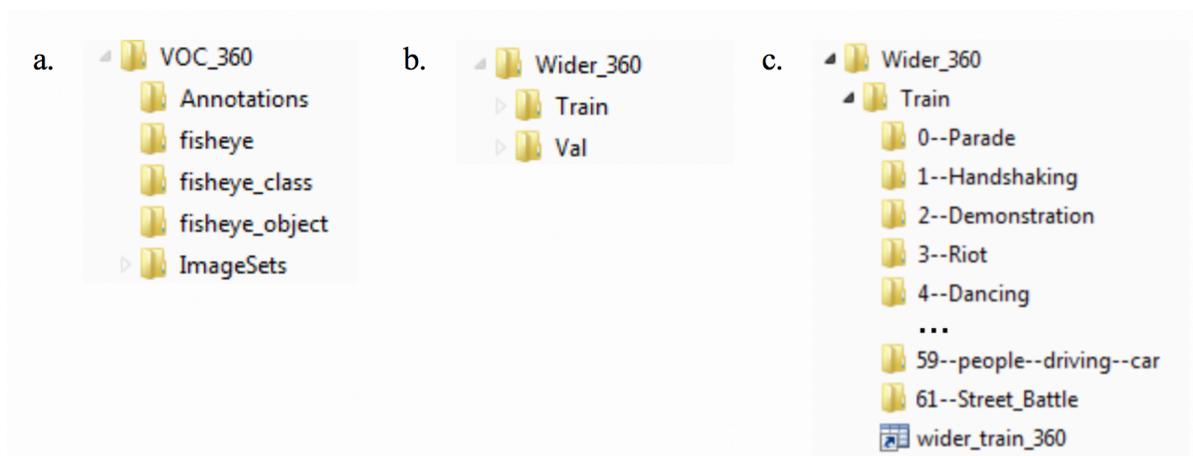

Fig. 1. Directory structures

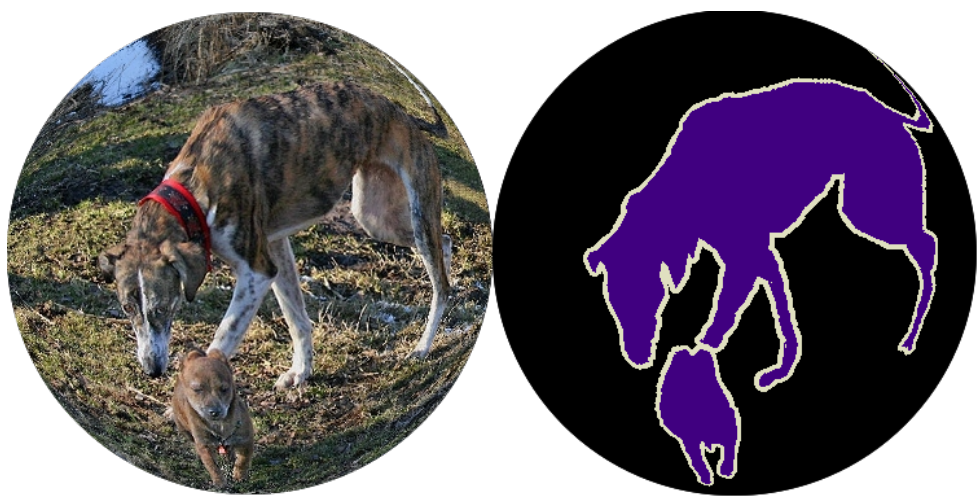

Fig. 2. Sample fisheye image with its segmentation mask from VOC-360

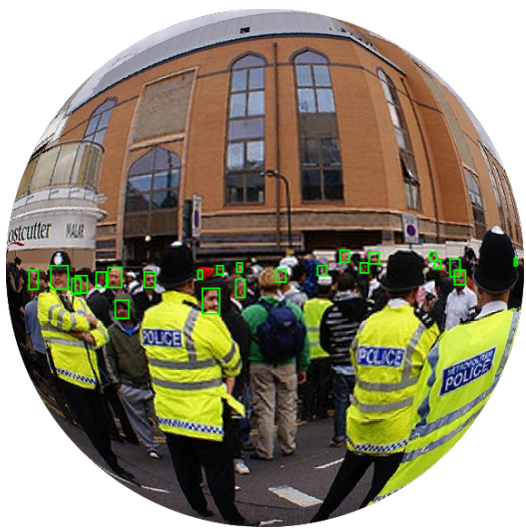

Fig. 3. Sample fisheye image with its ground-truth bounding boxes from Wider-360

2. **Experimental Design, Materials, and Methods**

The images were obtained from the existing public datasets – VOC2012 [1] and Wider Face [2] – and transformed into fisheye-looking images. The corresponding annotations were also converted into the fisheye image coordinate system. Square patches were sampled from the original images and converted to fisheye-looking images using the following generic transformation [3]:

$$(x', y') = \left( x \cdot \sqrt{1 - \frac{y^2}{2}}, y \cdot \sqrt{1 - \frac{x^2}{2}} \right) \quad (1)$$

$$(x'', y'') = \left( x' \cdot e^{-\frac{r^2}{4}}, y' \cdot e^{-\frac{r^2}{4}} \right) \quad (2)$$

Equation (1) converts the square patch to a circular patch. Here, $(x, y)$ are the normalized coordinates of the square patch, such that the center of the square patch is at $(0,0)$, while the four corners have the coordinates as $(\pm 1, \pm 1)$. The output coordinates $(x', y')$ refer to the produced circular patch. Equation (2) further squeezes the circular image towards the perimeter. Here, $r = \sqrt{(x')^2 + (y')^2}$ is the radial distance from the center of the circular patch, while the output coordinates $(x'', y'')$ refer to the final, fisheye-looking circular patch.

An alternative method to generate a fisheye-looking image from a rectangular or square image was proposed in [4], although no public data was provided with that work. The method in [4] is based on equidistant projection and requires the user to specify the focal length of the camera. In Fig. 4 below we show three images: a real fisheye image taken by the Ricoh Theta V 360-degree camera (left), an image generated from a square original by the proposed method (middle) and an image generated from the same original by the method from [4] (right) using one of the suggested focal length values ($f = 242$). The image generated by our method (middle) is a closer approximation to a real fisheye image (left).

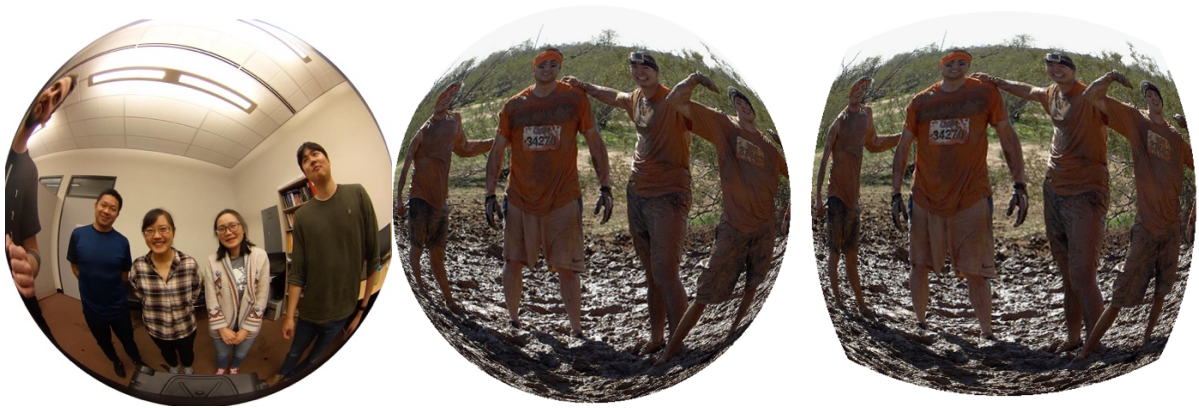

Fig. 4. Fisheye image from the Ricoh Theta V 360-degree camera (left), image generated by our proposed method (middle), and an image generated by the method from [4]

Each ground-truth annotation was converted to the new coordinate system. For segmentation annotations, the segmentation mask was treated as an image, and the transformation specified by equations (1)-(2) was applied to generate the segmentation mask for the fisheye image, as shown in the right part of Fig. 2. For bounding-box annotations, eight points were selected around the bounding box: four corner points and four edge midpoints, as depicted in Fig. 5. The same transformation specified by equations (1) and (2) was applied to these eight points to map them to the fisheye image coordinate system. The new points define a polygon in the fisheye image coordinate system, as shown in the right part of Fig. 5. The minimum axis-aligned bounding rectangle of these new eight points was found, shown in green in Fig. 5, and stored as the new bounding-box annotation for the fisheye image.

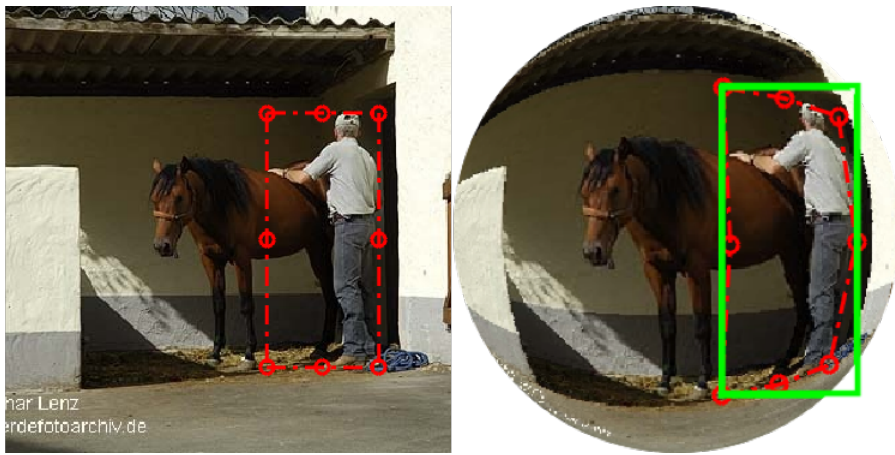

Fig. 5. Convert annotation box from square coordinate to circular coordinate

**Acknowledgments**
This work was supported in part by the Canadian Natural Sciences and Engineering Research Council (NSERC) Strategic Project Grant STPGP 494209.

**Specifications Table**

| | |
|---|---|
| Subject area | Computer vision, pattern recognition, machine learning |
| More specific subject area | Object classification, object detection, object recognition, object segmentation, face detection |
| Type of data | Images, annotations |
| How data was acquired | Data was created by processing images and annotations from two existing public datasets: VOC2012 and Wider Face. Images were converted from regular to fisheye via a nonlinear mapping, and the annotations were then mapped to the fisheye image coordinate system. Image and annotation conversion code was implemented in Matlab. |
| Data format | Raw data: JPG, XML, MAT files |
| Experimental factors | Indoor/outdoor scenes, variable illumination, various object types |
| Experimental features | The non-linear mapping used for conversion from regular to fisheye image is camera- and lens-independent, and captures the main characteristics geometric distortion found in fisheye images. |
| Data source location | Burnaby, BC, Canada. Simon Fraser University, School of Engineering Science. Latitude: 49.276765, Longitude: -122.917957 |
| Data accessibility | Public. <br> VOC-360: http://dx.doi.org/10.25314/ca0092b1-1e87-4928-b5f5-ebae30decb8d <br> Wider-360: https://researchdata.sfu.ca/pydio_public/c09804 |
| Related research article | This is a direct submission to Data in Brief, the most relevant research article from the reference list is [3]. |

**Value of the Data**

- VOC-360 is the first dataset for object detection in fisheye images. It will be useful to researchers and engineers in academia and industry to train object detectors directly on fisheye images.
- VOC-360 is also the first dataset for object segmentation in fisheye images. It will be useful to researchers and engineers in academia and industry to train vision-based AI systems for object segmentation on fisheye images.
- Wider-360 is the largest dataset for face detection in fisheye images. It will be useful to researchers and engineers in academia and industry to train face detectors directly on fisheye images.